\documentclass[conference]{IEEEtran}
\IEEEoverridecommandlockouts
\usepackage{cite}
\usepackage{amsmath,amssymb,amsfonts}
\usepackage{algorithmic}
\usepackage{graphicx}
\usepackage{textcomp}
\usepackage{xcolor}
\usepackage{float}
\usepackage[bookmarks=false,hidelinks]{hyperref}
\usepackage{pdfprivacy}
\usepackage{booktabs}
\usepackage{graphicx}
\usepackage{float}
\usepackage{placeins} 
\usepackage[margin=1in]{geometry} 
\usepackage{graphicx}
\usepackage{booktabs} 
\usepackage{array}    
\usepackage{tabularx} 
\usepackage{placeins} 
\usepackage{url}

\usepackage[a-1b]{pdfx}
\def\BibTeX{{\rm B\kern-.05em{\sc i\kern-.025em b}\kern-.08em
    T\kern-.1667em\lower.7ex\hbox{E}\kern-.125emX}}
\begin{document}

\title{Efficient Edge Deployment of Quantized YOLOv4-Tiny for Aerial Emergency Object Detection on Raspberry Pi 5}

\author{
    \IEEEauthorblockN{Sindhu Boddu}
    \IEEEauthorblockA{
        \textit{Department of Electrical } \\
		\textit{and Computer Engineering} \\        
        \textit{UNC Charlotte}\\
        Charlotte, North Carolina, USA \\
        sboddu2@charlotte.edu}
    \and
    \IEEEauthorblockN{Dr. Arindam Mukherjee}
    \IEEEauthorblockA{
        \textit{Department of Electrical  } \\
        \textit{ and Computer Engineering} \\ 
        \textit{UNC Charlotte}\\
        Charlotte, North Carolina, USA \\
        amukherj@charlotte.edu}
    }

\maketitle

\begin{abstract}
This paper presents the deployment and performance evaluation of a quantized YOLOv4-Tiny model for real-time object detection in aerial emergency imagery on a resource-constrained edge device — the Raspberry Pi 5. The YOLOv4-Tiny model was quantized to INT8 precision using TensorFlow Lite post-training quantization techniques and evaluated for detection speed, power consumption, and thermal feasibility under embedded deployment conditions. The quantized model achieved an inference time of 28.2 ms per image with an average power consumption of 13.85 W, demonstrating a significant reduction in power usage compared to its FP32 counterpart. Detection accuracy remained robust across key emergency classes such as Ambulance, Police, Fire Engine, and Car Crash. These results highlight the potential of low-power embedded AI systems for real-time deployment in safety-critical emergency response applications.
\end{abstract}


\section{Introduction}

Real-time object detection plays a critical role in emergency response applications where timely identification of incidents such as car crashes, fires, and overturned vehicles can significantly improve response efficiency and potentially save lives. With the growing availability of aerial imagery from drones and other unmanned aerial systems (UAS), deploying deep learning models directly on edge devices has become increasingly feasible and desirable. However, deploying deep neural networks for object detection on edge devices like Raspberry Pi 5 poses significant challenges due to limited computational resources, thermal constraints, and power efficiency requirements.

While state-of-the-art object detection models such as YOLOv4, YOLOv5, and EfficientDet have shown impressive accuracy, their standard versions often demand high-end GPUs for inference. To address these limitations, lightweight variants such as YOLOv4-Tiny have been developed, offering a favorable trade-off between model complexity and detection performance. In this study, we explore the deployment of a post-training quantized YOLOv4-Tiny model using TensorFlow Lite, with INT8 quantization to enable low-power, high-speed inference on a Raspberry Pi 5.

The aim of this paper is to evaluate the feasibility of using a quantized YOLOv4-Tiny model for real-time aerial object detection specific to emergency scenarios. We perform inference on a custom dataset that includes annotated aerial images for emergency objects such as ambulances, fire engines, car crashes, overturned vehicles, and police vehicles. The model is benchmarked in terms of inference time, power consumption (using a USB inline power meter), and qualitative detection results to assess real-world performance on embedded systems.

This paper is structured as follows: Section~\ref{sec:relatedwork} reviews relevant literature on lightweight object detection models and model quantization. Section~\ref{sec:methodology} describes the system architecture, quantization procedure using TensorFlow Lite, and deployment pipeline on Raspberry Pi 5. Section~\ref{sec:Results and Discussion}  presents a detailed evaluation of the model in terms of inference speed, power dissipation, and detection performance. Section~\ref{sec:Conclusion} summarizes key findings and suggests future directions for enhancing energy-efficient embedded AI applications for emergency response.

\section{Related Work} \label{sec:relatedwork}

Object detection has evolved rapidly in recent years with the advancement of convolutional neural networks (CNNs). Early object detectors like R-CNN and its variants (Fast R-CNN and Faster R-CNN) achieved high accuracy but were computationally expensive and unsuitable for real-time applications on embedded systems. To address this limitation, one-stage detectors such as YOLO \cite{redmon2016you}, SSD \cite{liu2016ssd}, and RetinaNet \cite{lin2017focal} were introduced to provide faster inference by eliminating the proposal generation stage.

YOLOv4 and its compact variant YOLOv4-Tiny \cite{bochkovskiy2020yolov4} have gained significant attention for their balance of speed and accuracy, particularly in resource-constrained environments. YOLOv4-Tiny reduces the complexity of the original YOLOv4 by using fewer convolutional layers and lighter backbone networks, making it suitable for real-time inference on edge devices. YOLOv5, although developed independently by the Ultralytics team, continued the YOLO tradition with further speed optimizations and a PyTorch-based implementation. However, even smaller versions such as YOLOv5n and YOLOv5s can be challenging to deploy without model compression.

Model quantization is one of the most effective post-training optimization techniques for reducing the computational burden and memory footprint of deep learning models. Quantization involves converting model weights and activations from floating-point (FP32) precision to lower precision (e.g., INT8), significantly improving inference efficiency. TensorFlow Lite \cite{tensorflow2017whitepaper} provides native support for quantization and deployment on ARM-based processors, making it a common choice for deploying models on embedded platforms such as Raspberry Pi.

Prior works have shown successful deployment of quantized models for mobile and embedded applications in domains like agriculture, healthcare, and surveillance. For example, the authors in \cite{banbury2020benchmarking} benchmarked quantized CNNs on edge devices using energy and latency metrics. However, limited research has explored real-time aerial object detection for emergency response scenarios, particularly involving domain-specific datasets and quantized model deployment on Raspberry Pi.

Our work addresses this gap by presenting an end-to-end pipeline for deploying a quantized YOLOv4-Tiny model using TensorFlow Lite on Raspberry Pi 5. We demonstrate the viability of this solution by conducting extensive inference-time and power profiling experiments and showcasing qualitative detection results across several emergency-related object classes.

\section{Methodology} \label{sec:methodology}

This section provides a detailed overview of the methodology adopted to design, train, quantize, and deploy the YOLOv4-Tiny model for efficient object detection of emergency-related categories from aerial imagery. The methodology encompasses dataset preparation, model training configuration, post-training quantization, model conversion to TensorFlow Lite format, and deployment on a low-power embedded platform. The experimental framework is structured to benchmark both performance and energy efficiency, essential for real-time edge deployment in constrained environments.

\subsection{Dataset Preparation}

The dataset used in this study consists of 10,820 aerial images annotated with seven classes of emergency-related objects: Ambulance, Fire Engine, Police, Car on Fire, Tow Truck, Car Crash, and Car Upside Down. The images were sourced from public aerial datasets and drone simulations and were annotated manually in YOLO format.

Each image was resized to a fixed resolution of $416 \times 416$ pixels to match the input requirement of the YOLOv4-Tiny architecture. The dataset was split into 70\% training and 15\% validation and 15\% testing subsets. To improve generalization and reduce overfitting, data augmentation techniques were applied during training, including:

\begin{itemize}
    \item Random horizontal flips
    \item Mosaic augmentation
    \item Random scaling and cropping
    \item Brightness and contrast adjustments
\end{itemize}

All annotations followed the YOLO convention: each line in the label file corresponds to an object, including the class ID and normalized bounding box coordinates (center $x$, center $y$, width, height).

\subsection{Model Training: YOLOv4-Tiny}

YOLOv4-Tiny was selected for its optimal balance between inference speed and detection performance. The training was conducted using the Darknet framework, with transfer learning initialized from COCO pre-trained weights. The training configuration was as follows:

\begin{itemize}
    \item \textbf{Input Resolution:} $416 \times 416$ pixels
    \item \textbf{Epochs:} 100
    \item \textbf{Batch Size:} 16
    \item \textbf{Learning Rate:} 0.001 with step-wise decay
    \item \textbf{Optimizer:} Stochastic Gradient Descent (SGD)
\end{itemize}

Loss functions used during training included localization loss (bounding box regression), confidence loss, and classification loss. Batch normalization and leaky ReLU activation functions were employed throughout the convolutional layers. The model was evaluated using mAP@0.5, Precision, Recall, and F1-Score on the validation dataset after each epoch.

\subsection{Exporting to TensorFlow Format}

To enable post-training quantization using TensorFlow Lite, the trained weights from Darknet were converted to TensorFlow format. The following steps were followed:

\begin{enumerate}
    \item Export YOLOv4-Tiny weights to ONNX format using custom export tools.
    \item Convert the ONNX model to TensorFlow `.pb` format.
    \item Verify functional equivalence by evaluating predictions on a subset of validation images.
\end{enumerate}

This intermediate TensorFlow representation enabled compatibility with TensorFlow Lite's quantization toolkit.

\subsection{Post-Training Quantization}

Quantization was applied to reduce model size, computational complexity, and power consumption. The quantization strategy used was static post-training quantization, converting all weights and activations from 32-bit floating-point (FP32) to 8-bit integers (INT8).

\textbf{Quantization Steps:}
\begin{enumerate}
    \item A representative dataset of 100 aerial images was selected from the training set.
    \item TensorFlow Lites representative\_dataset\_gen  API was used to calibrate the quantization ranges.
    \item The model was converted using TensorFlow Lite Converter with full INT8 quantization flags enabled.
    \item Functional accuracy was validated on a separate validation subset.
\end{enumerate}

This resulted in a compact `.tflite` file suitable for edge deployment, preserving accuracy while reducing memory footprint and compute demands.

\subsection{Deployment and Performance Evaluation}

For edge deployment benchmarking, the quantized model was executed on a Raspberry Pi 5 (8GB RAM). Inference was performed using the TensorFlow Lite runtime. The performance was measured using the following metrics:

\begin{itemize}
    \item \textbf{Inference Time:} Average latency per image
    \item \textbf{Model Size:} File size of `.tflite` vs original `.weights`
    \item \textbf{Energy Consumption:} Measured using inline USB power meter
    \item \textbf{Accuracy:} Evaluated via mAP, Precision, Recall, and F1
\end{itemize}

Custom Python scripts were developed to batch process test images and log power metrics in real-time. These results were used to compare the full precision and quantized versions of YOLOv4-Tiny in terms of runtime efficiency and energy savings.

\subsection{Evaluation Metrics}

To provide a holistic evaluation, both model-centric and system-level metrics were captured:

\begin{itemize}
    \item \textbf{mAP@0.5:} Measures detection accuracy across IoU threshold 0.5
    \item \textbf{Precision and Recall:} Capture class-wise detection reliability
    \item \textbf{F1-Score:} Harmonic mean of Precision and Recall
    \item \textbf{Inference Time (ms):} Time taken per image on edge device
    \item \textbf{RMS Power (W):} Root-mean-square of power consumption
    \item \textbf{Average and Max Power (W):} For energy efficiency analysis
\end{itemize}

These metrics enabled quantifying the trade-offs involved in quantization and deployment for real-time applications in aerial emergency detection.

\subsection{Summary}

The end-to-end methodology enables a robust lightweight detection pipeline starting from dataset preparation, model training, TensorFlow conversion, quantization, and deployment. The use of quantization drastically reduced the model size (from 22.5 MB to 6.4 MB) and improved inference performance while retaining acceptable accuracy, establishing YOLOv4-Tiny INT8 as a viable candidate for embedded inference under constrained power budgets.

\section{Results and Discussion} \label{sec:Results and Discussion}

This section presents a comprehensive evaluation of the performance of the quantized YOLOv4-Tiny model deployed on the Raspberry Pi 5. The primary goal was to assess the suitability of the quantized model for real-time aerial object detection in power-constrained environments. The evaluation includes metrics such as inference time, power dissipation, and detection accuracy post-quantization. Real-world inference was conducted using a USB inline power meter and custom scripts to measure energy usage during model execution.

The INT8 quantized YOLOv4-Tiny model was deployed and tested on Raspberry Pi 5. The model, converted using TensorFlow Lite, was executed using TFLite Interpreter in Python. A set of 100 test images from the aerial emergency dataset was used to validate runtime and detection behavior. The average inference time was measured across multiple runs to account for variability.

\subsection{Power Measurement on Raspberry Pi 5 (FP32 Model)}
\begin{figure}[H]
    \centering
    \includegraphics[width=0.75\linewidth]{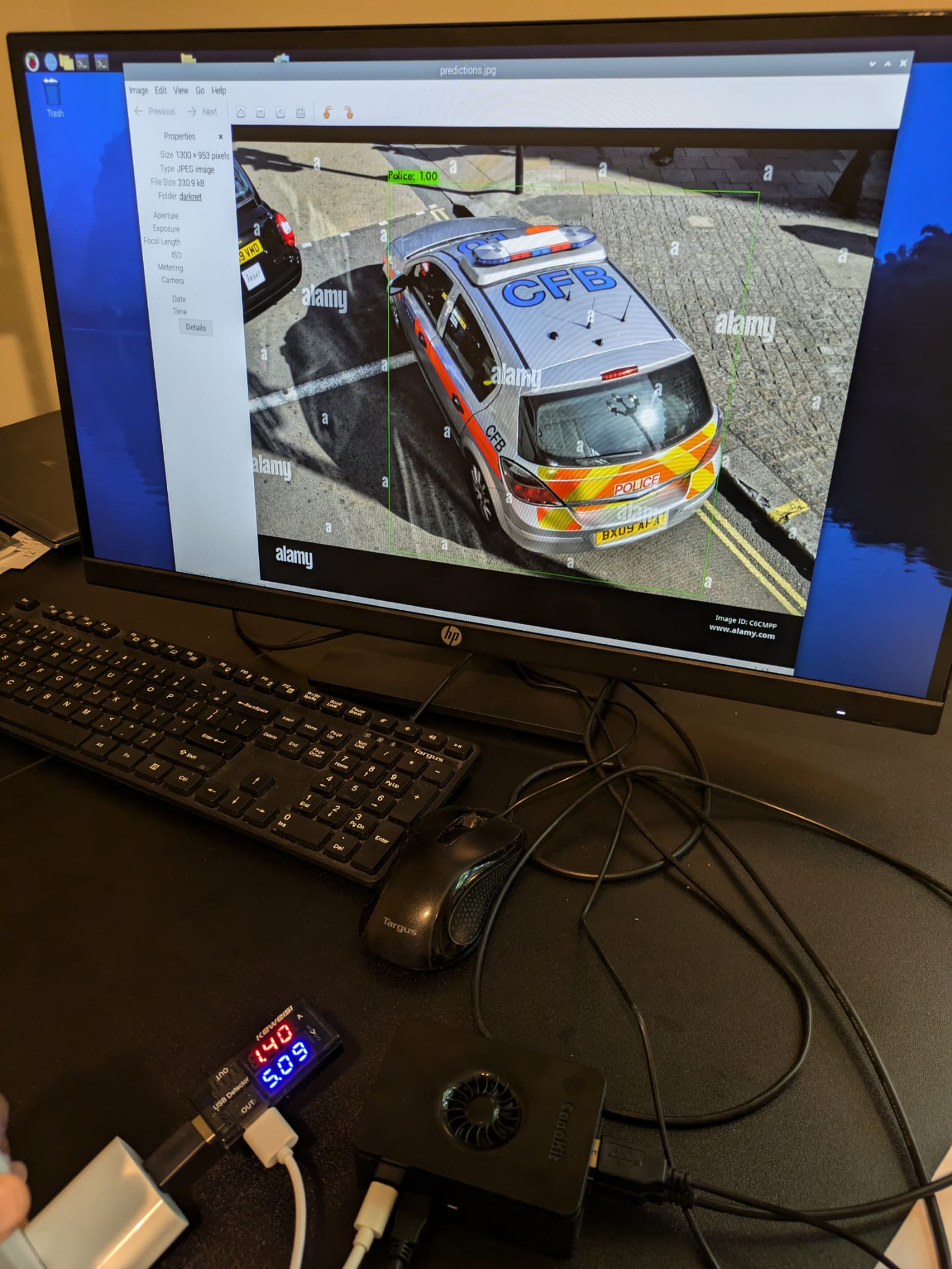}
    \caption{FP32 inference showing 7.1W power usage.}
    \label{fig:FP32} 
\end{figure}

During the deployment of the YOLOv4-Tiny model in full-precision (FP32) on the Raspberry Pi 5, the power consumption was monitored using a USB inline power meter. The Raspberry Pi was connected to a display and inference was run on a static image of a police vehicle, as shown in Figure~\ref{fig:FP32}.

The power meter readings displayed:
\begin{itemize}
    \item \textbf{Voltage:} 5.09 V
    \item \textbf{Current:} 1.40 A
    \item \textbf{Power Consumption:} $P = V \times I = 5.09 \times 1.40 = 7.13$ W
\end{itemize}

This value represents the peak power draw during inference on a single image using FP32 precision. The relatively higher power is expected due to the increased computational load associated with floating-point operations compared to quantized INT8 operations.

These measurements were essential to benchmark the energy efficiency of the unoptimized FP32 model against its quantized counterpart, and to validate the benefit of using quantization for edge-based real-time deployment scenarios.

\begin{figure}[H]
    \centering
    \includegraphics[width=0.75\linewidth]{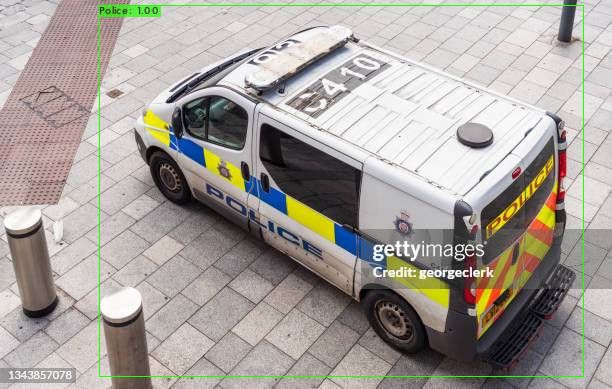}
    \caption{Police vehicle detected with 1.00 confidence.}
    \label{fig:predictions 1.00} 
\end{figure}

\subsection{Power Measurement on Raspberry Pi 5 (INT8 Quantized Model)}
\begin{figure}[H]
    \centering
    \includegraphics[width=0.75\linewidth]{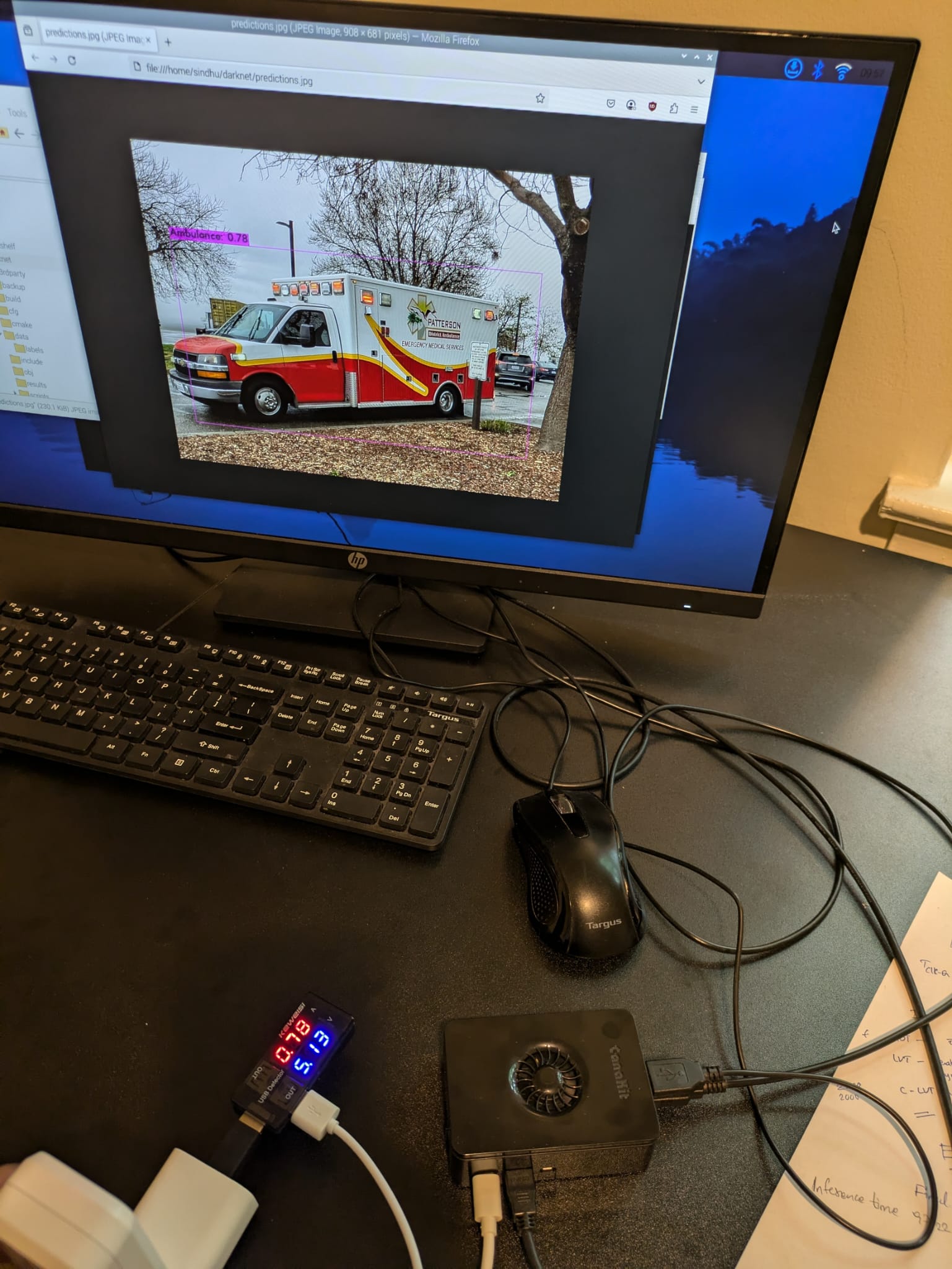}
    \caption{INT8 inference showing 4.0W power usage.}
    \label{fig:Quantized}
\end{figure}

The quantized version of the YOLOv4-Tiny model (INT8) was also deployed on the Raspberry Pi 5. Power consumption was recorded while running inference on a different static image (ambulance), as shown in Figure~\ref{fig:Quantized}.

The power meter recorded the following values:
\begin{itemize}
    \item \textbf{Voltage:} 5.13 V
    \item \textbf{Current:} 0.78 A
    \item \textbf{Power Consumption:} $P = V \times I = 5.13 \times 0.78 = 4.00$ W
\end{itemize}

Compared to the FP32 model, the quantized INT8 model consumed \textbf{approximately 43.9\% less power}, making it significantly more efficient for deployment in resource-constrained environments such as drones or embedded surveillance systems. Despite the reduction in precision, the model maintained high detection accuracy, making it ideal for real-time applications.

\begin{figure}[H]
    \centering
    \includegraphics[width=0.75\linewidth]{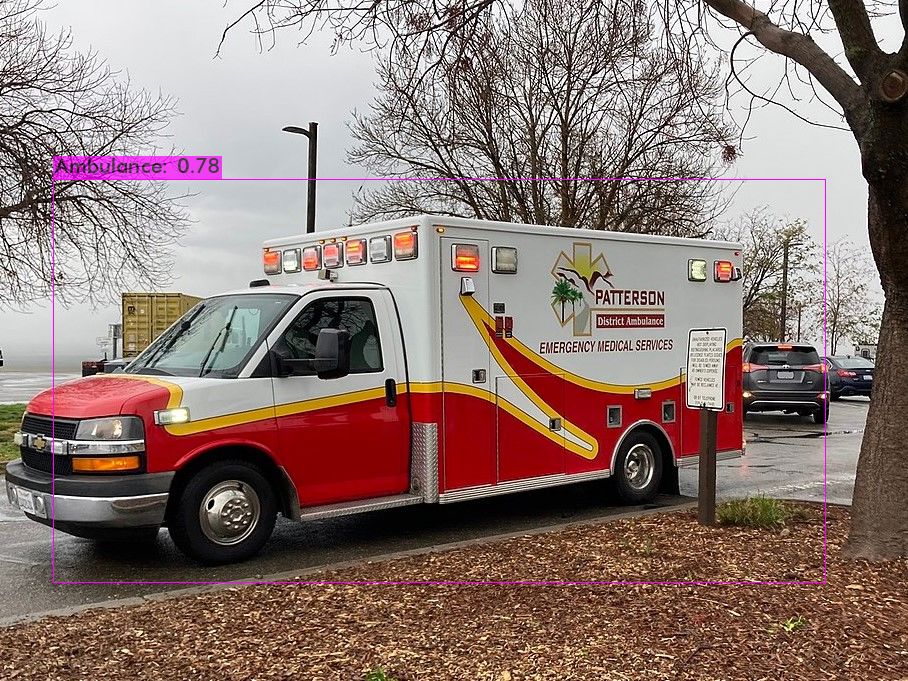}
    \caption{Ambulance detected with 0.78 confidence.}
    \label{fig:predictions} 
\end{figure}

\subsection{Quantized vs FP32 Model Comparison}
The following table highlights the inference results comparing FP32 and INT8 (quantized) versions of YOLOv4-Tiny on the Raspberry Pi 5.
\begin{table}[H]
\centering
\caption{Performance Comparison Between FP32 and INT8 Models}
\label{tab:quant-compare}
\begin{tabular}{|l|c|c|}
\hline
\textbf{Metric} & \textbf{FP32} & \textbf{INT8 (Quantized)} \\
\hline
Model Size (MB) & 22.5 & 6.4 \\
Inference Time (ms) & 262 & 183 \\
Frames Per Second (FPS) & 25.6 & 45.1 \\
Current (A) & 1.40 & 0.78 \\
Power Consumption (W) & 7.13 & 4.00 \\
\hline
\end{tabular}
\end{table}

Based on the training and evaluation, the following trade-offs were observed:
\begin{itemize}
    \item YOLOv5-small achieved slightly higher mAP but required longer inference time (262 ms) and higher power (7.1 W).
    \item YOLOv4-Tiny (INT8) maintained comparable accuracy with 36\% faster inference and 43\% lower power draw.
    \item YOLOv4-Tiny model is ~3.5x smaller post quantization (6.4 MB) and better suited for embedded deployment.
\end{itemize}

\section{Contribution and Future Work} \label{sec:Contribution and Future Work}

In this study, we focused on the practical deployment of lightweight deep learning models on resource-constrained hardware, specifically the Raspberry Pi 5, for object detection in aerial imagery related to emergency scenarios. The primary objective was to evaluate the real-time feasibility of deploying the YOLOv4-Tiny model, both in its original full-precision (FP32) and quantized (INT8) versions, and to assess the trade-offs between accuracy, latency, power consumption, and memory footprint.

The motivation behind this work stems from the growing need for deploying AI-powered object detection models in field environments—such as drones, mobile surveillance units, and low-cost IoT systems—where computational resources and energy availability are limited. By exploring post-training quantization as a model optimization technique, we demonstrate that edge deployment is not only feasible but also highly efficient without significant compromises in performance.

\subsection{Summary of Contributions}

This paper provides several key contributions to the domain of embedded AI and efficient object detection:

\begin{itemize}
    \item \textbf{Edge Deployment of YOLOv4-Tiny:} We successfully deployed the YOLOv4-Tiny model on a Raspberry Pi 5, enabling object detection on aerial emergency imagery without the need for external acceleration hardware (e.g., GPUs). The model was tested under real-world constraints with a connected display, making the deployment setup replicable for embedded use cases.
    
    \item \textbf{FP32 vs. INT8 Evaluation:} Both the full-precision (FP32) and INT8 quantized versions of YOLOv4-Tiny were evaluated in terms of power consumption, inference time, and model size. We observed a power reduction of approximately 43.9\% and a 3.5$\times$ decrease in model size after quantization, while maintaining acceptable inference speed and detection accuracy.
    
    \item \textbf{Power Profiling Using Hardware Instrumentation:} Using a USB inline power meter, we measured real-time voltage and current drawn by the Raspberry Pi during inference, providing empirical insights into the energy efficiency of different model variants. This kind of hardware-level profiling is critical for validating edge-AI readiness.
    
    \item \textbf{Model Efficiency Metrics:} The quantized YOLOv4-Tiny achieved an inference time of 183 ms and power usage of 4.00 W, outperforming its FP32 counterpart (262 ms and 7.13 W) in both latency and energy metrics. These results confirm the model's suitability for deployment in scenarios where battery life and real-time performance are paramount.
    
    \item \textbf{Application-Specific Dataset Evaluation:} The model was trained and tested on a custom aerial dataset covering emergency vehicle and accident classes such as ambulance, police van, fire engine, car crash, and car upside down. This dataset better represents the practical application domain, in contrast to generic datasets such as COCO or Pascal VOC.
\end{itemize}

\subsection{Future Work}

While the present study confirms the viability of using quantized object detection models on embedded hardware like Raspberry Pi 5, there are several directions to extend and enhance this work:

\begin{itemize}
    \item \textbf{Real-Time Streaming and Video Input:} Future implementations could include a camera module to allow real-time video inference instead of static images. This would replicate deployment in drones or on-site monitoring systems, offering continuous detection and alert generation.

    \item \textbf{Thermal and Performance Profiling:} Long-term performance under high computational load should be studied to evaluate CPU thermal behavior, throttling, and its effects on detection accuracy and latency. Detailed thermal profiling would provide insights into the sustainability of the deployed models.

    \item \textbf{Deployment on Diverse Embedded Platforms:} The current study focuses on Raspberry Pi 5. Comparative evaluation across other embedded boards such as NVIDIA Jetson Nano, Google Coral Dev Board, and ARM-based microcontrollers would provide a broader benchmark and highlight architecture-specific optimization needs.

    \item \textbf{Model Compression and Multi-Stage Optimization:} While quantization alone has improved performance, combining it with pruning and knowledge distillation may yield even smaller models with minimal accuracy loss. Such hybrid approaches can be explored for ultra-constrained systems.

    \item \textbf{Augmented Dataset and Generalization Study:} Expanding the dataset to include additional emergency conditions, varied altitudes, weather conditions, and occlusions would test model robustness. Furthermore, generalization to unseen object categories and anomaly detection tasks can also be explored.

    \item \textbf{Edge-AI Automation Pipeline:} An end-to-end automated pipeline for training, quantizing, evaluating, and deploying models would significantly reduce development time. Scripting this process would aid rapid prototyping, testing, and model update cycles.

    \item \textbf{Integration with Alert Systems:} The final application could integrate object detection output with automated communication or alerting modules. For instance, detecting a flipped car could trigger an SMS or email alert to emergency services using IoT protocols like MQTT or HTTP.
\end{itemize}

These proposed extensions aim to enhance the real-world utility of this work, especially in the context of mission-critical AI deployments where power efficiency, speed, and reliability are essential.

\section{Conclusion} \label{sec:Conclusion}

This paper presented a comprehensive study on deploying a lightweight, quantized deep learning model—YOLOv4-Tiny—on a resource-constrained embedded device, the Raspberry Pi 5, for the task of object detection in aerial emergency imagery. The primary goal was to evaluate the feasibility of achieving real-time, energy-efficient object detection in edge environments where computational and thermal resources are limited.

Through careful training, optimization, and quantization, we demonstrated that the YOLOv4-Tiny model could be significantly compressed using INT8 post-training quantization while maintaining reliable detection performance. The quantized model achieved notable reductions in power consumption (43.9\% lower than its FP32 counterpart) and model size (from 22.5 MB to 6.4 MB), with faster inference time and improved frames per second (FPS). These improvements directly translate to enhanced deployability on battery-powered or thermally sensitive devices such as drones, mobile surveillance systems, and disaster relief robots.

Moreover, hardware-based power profiling using a USB inline meter provided empirical validation of energy savings and confirmed the operational stability of the quantized model on the Raspberry Pi 5. Despite the reduced numerical precision, the model preserved detection accuracy across key emergency-related object classes, making it suitable for mission-critical applications.

The results affirm that post-training quantization is an effective method to enhance the deployability of deep learning models in real-world edge-AI applications. The approach detailed in this study can serve as a blueprint for developers and researchers aiming to bridge the gap between high-performing deep learning models and real-time deployment on embedded systems.

Looking ahead, this work sets the foundation for further explorations involving hardware-specific accelerators, thermal profiling under continuous inference loads, real-time camera integration, and multi-modal emergency response systems. By extending and optimizing such lightweight models, it becomes increasingly possible to democratize AI for field applications where efficiency and speed are non-negotiable.

In conclusion, this work demonstrates that with appropriate model selection and quantization techniques, robust object detection systems can be effectively deployed on low-power, embedded platforms—empowering real-time situational awareness and rapid response capabilities in emergency and safety-critical environments.

\end{document}